\documentclass{article}
\usepackage{arxiv}
\usepackage{cite}
\usepackage{amsmath,amssymb,amsfonts}
\usepackage{graphicx}
\usepackage{textcomp}
\usepackage{bm}
\usepackage{multirow}
\usepackage{xcolor}
\usepackage{subfigure}
\usepackage{algorithm}
\usepackage{algorithmicx}
\usepackage{algpseudocode}
\usepackage{amsmath}
\usepackage{graphics}
\usepackage{epsfig}
\usepackage{url}
\usepackage{makecell}
\DeclareMathOperator*{\argmin}{arg\,min}
\usepackage{xspace}
\newcommand{\T}{^{\textrm T}} 

\newcommand{\vct}[1]{\boldsymbol{#1}} 
\newcommand{\mat}[1]{\boldsymbol{#1}} 
\newcommand{\ie}{\emph{i.e.}\xspace} 
\newcommand{\etal}{\emph{et al.}\xspace} 

\begin{document}

\title{Meta Ordinal Regression Forest For Learning with Unsure Lung Nodules}

\author{\name Yiming Lei \email{ymlei17@fudan.edu.cn} \\
\name Haiping Zhu \email{hpzhu14@fudan.edu.cn} \\
\name Junping Zhang \email{jpzhang@fudan.edu.cn} \\
\addr{Shanghai Key Laboratory of Intelligent Information Processing, School of Computer Science, Fudan University, Shanghai 200433, China}\\
\name Hongming Shan
\email{hmshan@fudan.edu.cn} \\
\addr{Institute of Science and Technology for Brain-inspired Intelligence and  MOE Frontiers Center for Brain Science, Fudan University, Shanghai 200433, China} \\
\addr{Shanghai Center for Brain Science and Brain-inspired Technology, Shanghai 201210, China}
}

\maketitle

\begin{abstract}
Deep learning-based methods have achieved promising performance in early detection and classification of lung nodules, most of which discard unsure nodules and simply deal with a binary classification---\emph{malignant} vs \emph{benign}. Recently, an unsure data model (UDM) was proposed to incorporate those unsure nodules by formulating this problem as an ordinal regression, showing better performance over traditional binary classification. To further explore the ordinal relationship for lung nodule classification, this paper proposes a meta ordinal regression forest (MORF), which improves upon the state-of-the-art ordinal regression method, deep ordinal regression forest (DORF), in three major ways. First, MORF can alleviate the biases of the predictions by making full use of deep features while DORF needs to fix the composition of decision trees before training. Second, MORF has a novel grouped feature selection (GFS) module to re-sample the split nodes of decision trees. Last, combined with GFS, MORF is equipped with a meta learning-based weighting scheme to map the features selected by GFS to tree-wise weights while DORF assigns equal weights for all trees. Experimental results on LIDC-IDRI dataset demonstrate superior performance over existing methods, including the state-of-the-art DORF.
\end{abstract}

\section{Introduction}
\label{sec:introduction}
Early detection and classification of lung nodules are critical for reducing the high mortality of lung cancer. With the rapid development of deep learning, deep neural networks (DNNs)-based computer-aided diagnosis of lung cancer has achieved exciting progress in recent years. According to the malignancy progression of lung nodules, a nodule in the public LIDC-IDRI dataset~\cite{LIDC} is labeled with malignancy scores from 1 to 5 representing the increase of the malignant degree. In LIDC-IDRI, there are 1,108 benign nodules of average score less than 2.5,  1,007 unsure nodules of average score between 2.5 and 3.5, and 510 malignant nodules of average score greater than 3.5. That is, a large number of nodules are unsure and then discarded in the conventional binary classification methods~\cite{fully3d2017,2017multilevel3d,lei2020shape}. In other words, the unsure nodules are between benign and malignant, and cannot be determined by radiologists based on current scans~\cite{wu2019learning}. Considering that deep learning methods are data-hungry in nature, \emph{how to leverage those unsure nodules for training a deep learning model} is an important topic. 

In this paper, we regard the problem of learning with unsure data as an ordinal regression problem~\cite{gutierrez2015ordinal}, which predicts the ordinal variable of malignancy: benign, unsure, and malignant. Recently, some ordinal regression methods~\cite{wu2019learning,liu2018ordinal,beckham2017unimodal} focus on solving this problem without consideration of applying the ordinal relationship in the training strategy.

In recent years, a series of ordinal regression methods based on random forest and convolutional neural network (CNN) have been successfully used for human age estimation through facial images~\cite{shen2018deep,zhu2020deep}. To utilize the global ordinal relationship, Zhu~\etal proposed the deep ordinal regression forest (DORF) and achieved the state-of-the-art results in age estimation~\cite{zhu2020deep}. Specifically, DORF uses the feature vector generated by the fully-connected (FC) layer to construct a random forest that contains many decision trees, and the final results are the average of all leaf nodes. There are two drawbacks of DORF. First, the split nodes in each tree are randomly selected from the FC feature, and the selected nodes are fixed during training. Consequently, all the elements in the feature vector of the FC layer are not fully explored. Second, all the trees are assigned with the same weights during training; \ie, each tree makes an equal contribution to the final prediction. These drawbacks result in the prediction biases with respect to different trees (tree-bias).

\begin{figure}[t]
\includegraphics[width=1.0\linewidth]{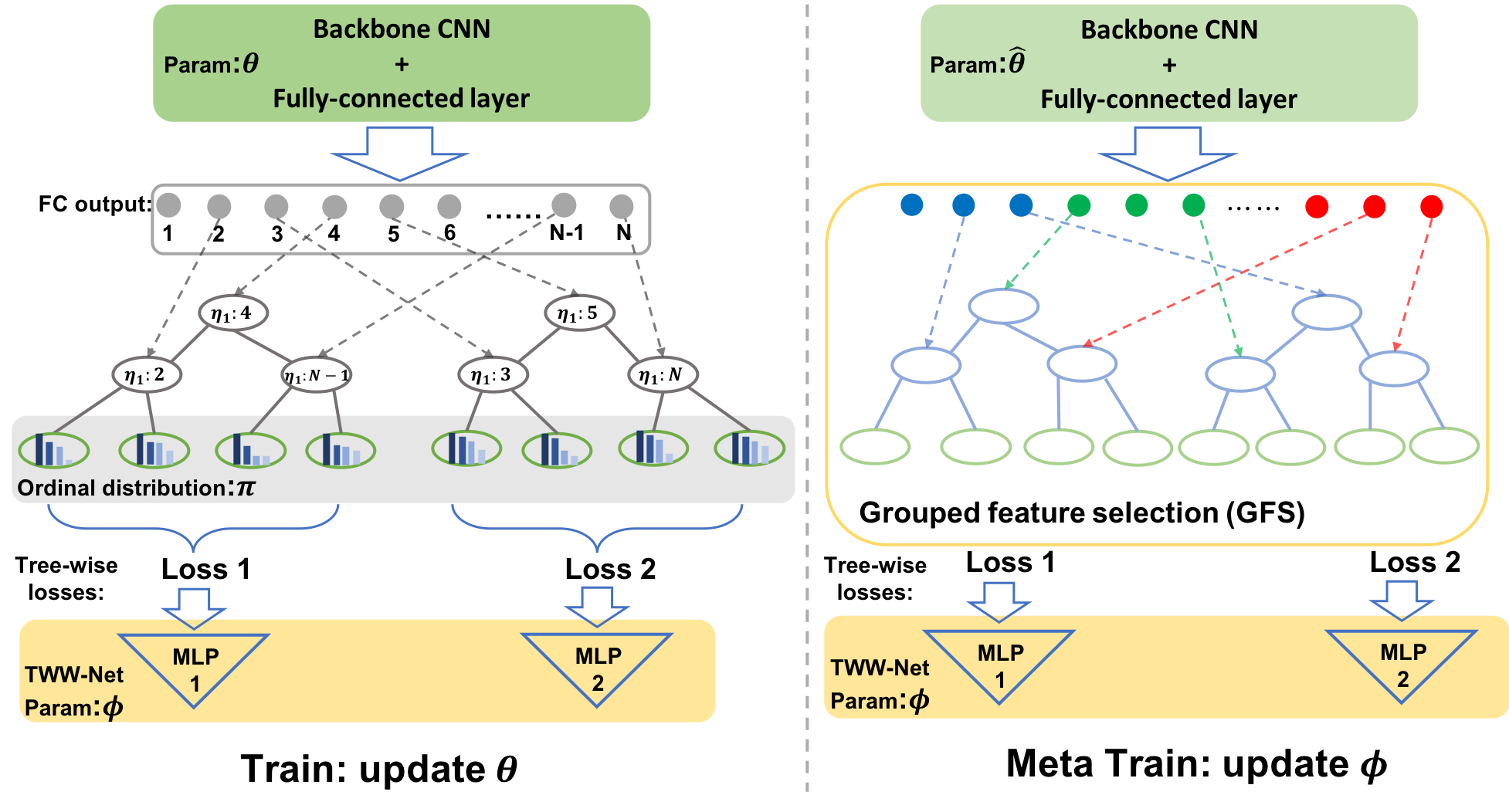}
\caption{The proposed MORF framework. On the left side is deep ordinal regression forest with random selection of trees followed by the TWW-Net. On the right side is deep ordinal regression forest with GFS followed by the TWW-Net. The right side only works in meta training stage that uses GFS features to guide the update of $\vct{\theta}$. The total framework includes three parts of parameters: $\vct{\theta}$ (CNN), $\vct{\pi}$ (leaf nodes), and $\vct{\phi}$ (TWW-Net).
}
\label{fig:framework}
\end{figure}

To address these drawbacks, we propose a meta ordinal regression forest (MORF) to make full use of the FC feature and alleviate the prediction bias. More specifically, we design a tree-wise weighting network (TWW-Net) to assign each tree with a specific weight, which is learned during training. The input to the TWW-Net is the loss generated by training/meta data. Here, meta data are obtained through the proposed grouped feature selection (GFS) module, which is used to sort the output of the FC layer from large to small, and then split the sorted elements into $\mathcal{N}$ groups, where $\mathcal{N}$ corresponds to the number of split nodes in one tree; see right subfigure in Fig.~\ref{fig:framework}. As a result, all the trees locally select the elements for themselves as the new split nodes. It should be noted that this re-sampling module only works in the meta training stage and has no impact on the structure of the forest at the inference stage. Hence, the gradient generated by training loss can be guided and regularized by the meta loss. In addition to that,  the predictions of all trees in MORF are in line with the final predictions. We also highlight that this merit does not exist in current deep regression forest methods~\cite{shen2018deep,zhu2020deep}. 

Our meta training algorithm is similar to~\cite{shu2019meta} but differs in the construction of the meta dataset. The construction method in~\cite{shu2019meta} is not applicable in the medical domain that some fine-grained classes contain fewer samples, and this may trigger a severe class imbalance if we pick out a part of samples to make meta dataset. We instead use the GFS module to construct the meta dataset.

The contributions are summarized as follows. 1) We propose a meta ordinal regression forest (MORF) for lung nodule classification with unsure nodules. The MORF contains a GFS module to improve the efficiency of feature reuse. 2) The proposed TWW-Net addresses the tree-bias problem by assigning each tree with a specific meta learned weight. 3) Experimental results demonstrate the effectiveness of TWW-Net weighting scheme. Our MORF achieves the best performance compared with other methods.

\section{Methods}
\label{sec:methods}
In this section, we introduce our MORF framework comprising a CNN backbone, a differentiable random forest, and the meta weighting scheme in conjunction with the GFS module and TWW-Net.

\subsection{Ordinal Regression Forest (ORF)}
Ordinal regression aims to learn a mapping function $h:\mathcal{X} \to \mathcal{Y}$, where $\mathcal{X}$ represents the input set of lung nodules and $\mathcal{Y} = \{r_{1}, r_{2}, \ldots, r_{K}\}$ is the ordinal label space such that $r_{1} \preceq r_{2} \preceq \ldots \preceq r_{K}$. In this study, $K=3$ and the $r_{1}$, $r_{2}$ and $r_{3}$ denote benign, unsure, and malignant, respectively. Then the label $y$ can be converted to an ordinal distribution label $\vct{o} = (o^{1}, o^{2}, \ldots, o^{K-1})\T \in \mathcal{O}$~\cite{zhu2020deep}, where $o^{k} = 1$ if $y>r_{k}$, otherwise $o^{k} = 0$. In order to guarantee the monotonically decreasing property, $\vct{o}$ should satisfy the constraint $o^{1} \ge o^{2} \ge \ldots \ge o^{K-1}$. The ordinal label of a sample $\vct{x}$ is given by the leaf node in ORF, and the probability of the sample $\vct{x}$ falling into the $l$-th leaf node is defined as: 
\begin{equation}
p(l | \vct{x}; \mat{\theta}) = \prod_{n \in \mathcal{N}} s_{n}(\vct{x}; \mat{\theta})^{\vct{1}[l \in \mathcal{L}_{n}^{l}]} \left(1 -  s_{n}(\vct{x}; \mat{\theta})^{\vct{1}[l \in \mathcal{L}_{n}^{r}]}\right),
\label{falling}
\end{equation}
where $\mathcal{L}_{n}^{l}$ and $\mathcal{L}_{n}^{r}$ represent the subsets of leaf nodes belonging to left and right sub-trees of split node $n$, and $s_{n}(\vct{x};\mat{\theta}) = \sigma(f_{\eta(n)}(\vct{x};\mat{\theta}))$ is the split function that determines which node (left or right) a sample should be assigned to, where $\sigma(\cdot)$, $f(\cdot; \mat{\theta})$ and $\eta(n)$ denote the sigmoid function, backbone network and index function, respectively. And $\eta(n)$ constructs the corresponding relationship between the $n$-th split node and the $\eta(n)$-th element of the output of $f$~\cite{shen2018deep,zhu2020deep}.

Next, the output of one tree is the average of the outputs of all leaf nodes, which can be defined as mapping $\vct{g} : \mathcal{X} \to \mathcal{O}$:
\begin{equation}
\vct{g}(\vct{x}; \mat{\theta}, \mat{\mathcal{T}}) = \sum_{l \in \mathcal{L}} p(l | \vct{x}; \mat{\theta}) \vct{\pi}_{l},
\label{orf}
\end{equation}
where $\mat{\mathcal{T}}$ denotes one decision tree, and $\mathcal{L}$ the set of leaf nodes. The term $\vct{\pi}_{l}$ in Eq.~\eqref{orf}  contains the ordinal distribution of the leaf node $l$; \ie, $\vct{\pi}_l = (\pi_{l}^{1}, \pi_{l}^{2}, \ldots, \pi_{l}^{K-1})\T$. $\vct{\pi}$ can be updated jointly with the back propagation of the DNN via solving a constrained optimization problem~\cite{shen2018deep,zhu2020deep}. Finally, the output of the forest is the average of all trees:
\begin{equation}
P(\vct{x}) = \frac{1}{T} \sum_{t = 1}^{T} \vct{g}(\vct{x}; \mat{\theta}, \mat{\mathcal{T}}_t),
\end{equation}
where $T$ is the total number of trees. We can see that each tree has the same weight for the final prediction.

\subsection{Grouped Feature Selection}
The traditional deep forest construction does not make full use of the FC feature, which results in the tree-wise prediction fluctuations. In order to address this problem, we propose a grouped feature selection (GFS) module to simply rank all the elements in the FC feature vector, and then split the ranked elements into $\mathcal{N}$ groups, which corresponds to the number of split nodes in one tree. For example in Fig.~\ref{fig:framework}, we can see that there are three colored groups, and each node in one tree randomly selects the feature from one group. Hence, the feature vector can be locally re-sampled for one node, and the split nodes of one tree can globally cover all FC features.

\subsection{Meta Training}
\label{subsec:meta_train}
As for the existing ORF algorithms being inferior in balancing the contributions of each tree which is caused by its own prediction bias, we propose a TWW-Net to assign each tree with a specific weight while reducing the tree-bias. In practice, the TWW-Net is formulated as several MLPs ($V_{t}$) due to its universal approximation~\cite{shu2019meta}, each of which corresponds to one tree. In Fig.~\ref{fig:framework}, we can see that the MORF includes three parts of parameters: the ordinal distribution of leaf nodes $\vct{\pi}$, the backbone parameter $\vct{\theta}$, and the TWW-Net parameter $\vct{\phi}$. Note that $\vct{\pi}$ can be  updated according to~\cite{zhu2020deep}. The backbone parameter $\vct{\theta}$ and the TWW-Net parameter $\vct{\phi}$ are learned via optimizing the following objective:
\begin{align}
 \vct{\theta^{*}}(\vct{\phi}) = \argmin_{\vct{\theta}} L^{\mathrm{tr}}(\vct{\theta}; \vct{\phi})  = \frac{1}{N}\sum_{i=1}^{N} \frac{1}{T}\sum_{t=1}^{T} V_{t}^{i}\left[R_{t}^{i}(\vct{\theta}, \vct{\pi}; S^{\mathrm{tr}}); \vct{\phi}\right] \cdot R_{t}^{i}(\vct{\theta}, \vct{\pi}; S^{\mathrm{tr}}),\label{eq:theta_obj} 
\end{align}
where $N$, $S^{\mathrm{tr}}$, and $R_{t}$ denote the number of training images, the training set, and the classification loss of the $t$-th tree calculated through the results of Eq.~\eqref{orf}, respectively. Note that the input of the weight generation net $V_{t}$ is the corresponding loss $R_{t}$. In summary, Eq.~\eqref{eq:theta_obj} imposes the specific weight on each tree via corresponding $V_{t}$.

For the TWW-Net, the parameter $\vct{\phi}$ is learned from the following objective:
\begin{align}
 \vct{\phi^{*}} &= \argmin_{\vct{\phi}} L^{\mathrm{meta}}(\vct{\theta^{*}}(\vct{\phi})) 
              = \frac{1}{M} \sum_{j=1}^{M} \frac{1}{T} \sum_{t=1}^{T} R_{t}^{j}(\vct{\theta^{*}}(\vct{\phi}), \vct{\pi}; S^{\mathrm{meta}}),\label{eq:phi_obj}
\end{align}
where $M$ is the number of meta data. This objective indicates that $\vct{\phi}$ is updated based on the optimal backbone parameter $\vct{\theta^{*}}$. In order to learn the $\vct{\theta}$ and $\vct{\phi}$ in one optimization loop, we adopted the meta learning manner which requires the second order derivatives (Hessian matrix) to update $\vct{\phi}$~\cite{shu2019meta}. Note that, when the model ($\vct{\theta}$) is well trained, the TWW-Net ($\vct{\phi}$) will be removed at inference stage.

\section{Experimental Setup and Results}
\label{sec:experiments}

\subsection{Data Preparation}
\label{subsec:data}
LIDC-IDRI is a publicly available dataset for LDCT-based pulmonary nodule analysis, which includes 1,010 patients who underwent chest CT. Each nodule was rated from 1 to 5 by four thoracic radiologists, indicating an increasing probability of malignancy. In this paper, the ROI of each nodule was obtained at its annotated center, with a square shape of a doubled equivalent diameter. An averaged score of a nodule was used as ground-truth for model training. All volumes were resampled to have $1$mm spacing (original spacing ranged from $0.6$mm to $3.0$mm) in each dimension, then the training and testing data were obtained by cropping the $32 \times 32 \times 32$ volumes according to the annotated centers. 

\subsection{Implementation Details}
\label{subsec:implement}
In our implementation, we chose three widely used networks as the backbone including ResNet-18, ResNet-34, and VGG-16~\cite{ResNet,vgg}. Due to the limited training data of LIDC-IDRI dataset, we adopt the 2D version of the chosen backbones~\cite{lei2020shape}. Therefore, the depth of the 3D input volume is viewed as the channel of the 2D patch. In other words, the input nodule has $32$ channels when training with pre-trained backbones. For the FC layer at the end of the backbone, the number of output units may not be equal to the number of classes; it is set as $256$ in our MORF and DORF. 

The hyperparameters for all experiments are set as follows: the learning rate is $0.001$ and decayed by $0.1$ every $80$ epochs; the size of a mini-batch size is $16$, and weight decay for Adam optimizer is $0.0001$~\cite{kingma2014adam}. The loss functions used in normal $n$-class classification (such as ResNet), MORF, and DORF, and each tree-wise loss during meta training are the standard cross-entropy (CE) loss.

\subsection{Performance Comparison}
In this section, we compare the results of our MORF and other ordinal regression methods when training with unsure data. Here, we focus more on the recall of the malignant class and the precision of the benign class, which is consistent with the conservative-aggressive strategy~\cite{wu2019learning}. All the experimental results are obtained through 5-fold cross validation.

In Table~\ref{tab:train3_test3}, we evaluate the performances of related methods by using three different backbone networks. The MORF achieves the highest accuracy in all settings compared with other methods. ResNet-18- and ResNet-34-based MORFs also perform better in the precision of benign, recall of malignant, and corresponding F1 scores. This indicates that there are fewer false positives of benign and fewer false negatives of malignant, which encourages more follow-ups for nodules with a score between 2.5 and 3.5 and alleviates the loss of missing diagnosis. On the other hand, although the VGG-based MORF achieves the best accuracy and precision of benign, it is weaker on recalls of both malignant and benign. And it obtains the highest precision of malignant indicating that fewer benign or unsure samples are misdiagnosed as malignant. More importantly, it remains the best recall and F1 score of the unsure class, which recommends the more accurate diagnosis of follow-ups.

Another superiority of our MORF is that it performs better in the recall of unsure than others. This indicates the clinical significance on two sides: 1) Fewer unsure samples can be misdiagnosed as benign (needs follow-up), which avoids the missing diagnosis probability of unsure samples to be developed into malignant. 2) Fewer unsure samples can be misdiagnosed as malignant which presents the importance of reducing misdiagnosis. This merit is not discussed in~\cite{wu2019learning}, and the recall of unsure using MORF is superior to that of UDM.

\begin{table*}[t]
\tiny
\caption{Performance comparison on LIDC-IDRI dataset. Following~~\cite{wu2019learning}, the values with underlines indicate the best results while less important in the clinical diagnosis.}
\renewcommand{\arraystretch}{1.3}
\setlength{\tabcolsep}{5pt}
\centering
\begin{tabular}{|c|l|*{10}{c|}}
\hline
\multirow{2}{*}{Backbone} & \multirow{2}{*}{Method} & \multirow{2}{*}{Accuracy} & \multicolumn{3}{c|}{Benign} & \multicolumn{3}{c|}{Malignant} & \multicolumn{3}{c|}{Unsure} \\ \cline{4-12}
 &  &  & Precision & Recall & F1 score & Precision & Recall & F1 score & Precision & Recall & F1 score \\
\hline
\hline
\multirow{6}*{\makecell{ResNet-18}}
& CE Loss & 0.542 & 0.544 & 0.722 & 0.620 & 0.586 & 0.644 & 0.613 & 0.496 &  0.290 & 0.366 \\
& Poisson~\cite{beckham2017unimodal} & 0.527 & 0.536 & 0.605 & 0.568 & 0.590 & 0.584 & 0.587 & 0.477 & 0.410 & 0.441 \\
& NSB~\cite{liu2018ordinal} & 0.534 & 0.517 & \underline{0.807} & 0.630 & 0.607 & 0.673 & 0.638 & 0.500 & 0.160 & 0.242 \\
& UDM~\cite{wu2019learning} & 0.546 & 0.553 & 0.767 & 0.643 & 0.581 & 0.495 & 0.535 & 0.504 & 0.325 & 0.395 \\
& DORF~\cite{zhu2020deep} & 0.568 & 0.569 & 0.713 & 0.633 & 0.632 & 0.613 & 0.623 & \underline{0.523} & 0.385 & 0.443 \\
& \textbf{MORF (ours)} & \textbf{0.573} & \textbf{0.695} & 0.623 & \textbf{0.657} & \underline{0.627} & \textbf{0.683} & \textbf{0.654} & 0.479 & \textbf{0.685} & \textbf{0.564} \\
\hline
\hline
\multirow{6}*{\makecell{ResNet-34}}
& CE Loss & 0.534 & 0.540 & \underline{0.722} & 0.618 & 0.625 & 0.644 & 0.634 & 0.443 & 0.270 & 0.335 \\
& Poisson~\cite{beckham2017unimodal} & 0.540 & 0.577 & 0.502 & 0.537 & \underline{0.714} & 0.545 & 0.618 & 0.458 & 0.520 & 0.512 \\
& NSB~\cite{liu2018ordinal} & 0.544 & 0.580 & 0.619 & 0.599 & 0.608 & 0.614 & 0.611 & 0.462 & 0.425 & 0.443 \\
& UDM~\cite{wu2019learning} & 0.536 & 0.537 & 0.659 & 0.592 & 0.578 & 0.554 & 0.566 & \underline{0.510} &  0.390 & 0.442 \\
& DORF~\cite{zhu2020deep} & 0.542 & 0.581 & 0.605 & 0.593 & 0.616 & 0.524 & 0.566 & 0.466 & 0.480 & 0.473 \\
& \textbf{MORF (ours)} & \textbf{0.586} & \textbf{0.619} & 0.619 & \textbf{0.619} & 0.504 & \textbf{0.700} & \textbf{0.656} & 0.509 & \textbf{0.560} & \textbf{0.533} \\
\hline
\hline
\multirow{6}*{\makecell{VGG-16}}
& CE Loss & 0.517 & 0.538 & 0.668 & 0.596 & 0.562 & 0.495 & 0.526 & 0.456 & 0.360 & 0.402 \\
& Poisson~\cite{beckham2017unimodal} & 0.542 & 0.548 & \underline{0.794} & \textbf{0.648} & 0.568 & \textbf{0.624} & 0.594 & 0.489 & 0.220 & 0.303 \\
& NSB~\cite{liu2018ordinal} & 0.553 & 0.565 & 0.641 & 0.601 & 0.566 & 0.594 & 0.580 & \underline{0.527} & 0.435 & 0.476 \\
& UDM~\cite{wu2019learning} & 0.548 & 0.541 & 0.767 & 0.635 & 0.712 & 0.515 & 0.598 & 0.474 & 0.320 & 0.382 \\
& DORF~\cite{zhu2020deep} & 0.559 & 0.590 & 0.627 & 0.608 & 0.704 & 0.495 & 0.581 & 0.476 & 0.515 & 0.495 \\
& \textbf{MORF (ours)} & \textbf{0.580} & \textbf{0.660} & 0.479 & 0.556 & \underline{0.719} & 0.596 & \textbf{0.652} & 0.492 & \textbf{0.680} & \textbf{0.571} \\
\hline
\end{tabular}
\label{tab:train3_test3}
\end{table*}

\subsection{Effectiveness of Weighting Scheme of TWW-Net}

\begin{figure}[t]
\centering
 \includegraphics[width=1.0\linewidth]{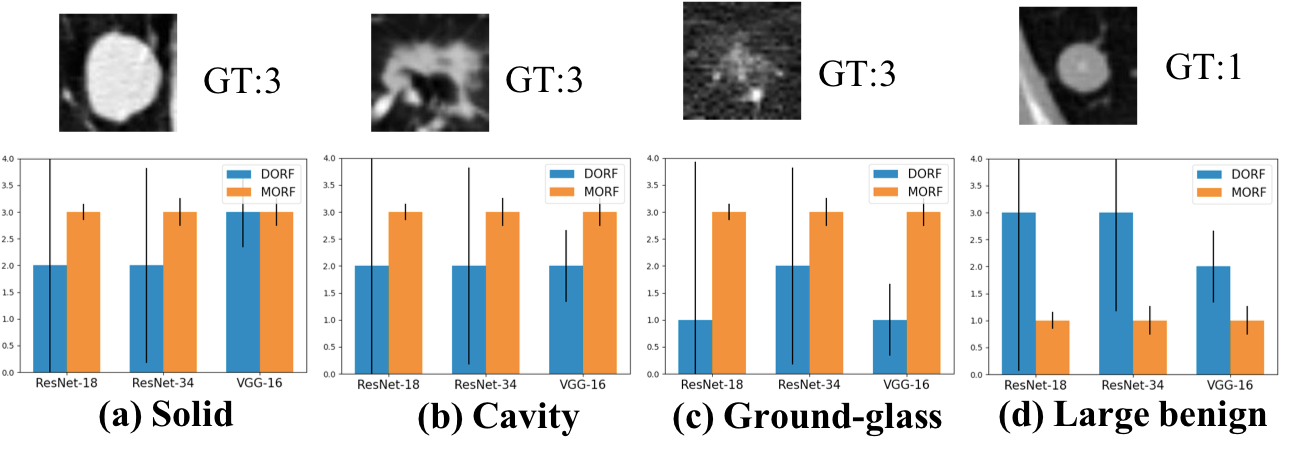}
\caption{Some prediction results of MORF and DORF. The $y$ axis represents the prediction results: 1, 2 and 3 represent benign, unsure and malignant. The GT denotes ground truth. The black vertical line on each bar represents the variance over the predictions of all trees.}
\label{fig:var}
\end{figure}
In order to demonstrate the effectiveness of TWW-Net weighting scheme, we estimate the variance over predictions of all trees: \begin{align}
\mathrm{var} := \frac{1}{T} \sum_{t=1}^{T} |p_{t} - \hat{p}|^{2},
\end{align}
where $p_{t}$ is the prediction of $t$-th tree and $\hat{p}$ is the final prediction, \ie average result of all trees. In Fig.~\ref{fig:var}, we compare the predictions and the corresponding $\mathrm{var}$s on some special nodules. We can see that the MORF predicted more accurately than DORF on these nodules when using different backbones. Especially for (\textbf{c}), the DORF makes the wrong predictions, which could result in a severe medical loss to the patients. For the nodules (\textbf{a}) and (\textbf{d}), the DORF is prone to regarding them as unsure and this is due to the less efficiency of feature reuse in DORF. It is obvious that the $\mathrm{var}$s of MORF are much lower than those of DORF which indicates that all trees in MORF predict consistently, and this benefits from the guidance of our GFS and meta training procedure.

\section{Conclusion}
\label{sec:con}
In this paper, we proposed a meta learning-based ordinal regression forest, termed MORF, and a GFS meta data construction module. The MORF exhibits better classification performance compared with state-of-the-art ordinal regression. The overall performance of MORF implies the reduction of loss of missing diagnosis and the misdiagnosis. We also provide an analysis of our meta weighting scheme and validate the prediction stability via the prediction variances over all trees.


\begin{thebibliography}{10}

\bibitem{LIDC}
Samuel~G Armato~III, Geoffrey McLennan, Luc Bidaut, et~al.
\newblock The lung image database consortium ({LIDC}) and image database
  resource initiative ({IDRI}): {A} completed reference database of lung
  nodules on {CT} scans.
\newblock {\em Medical Physics}, 38(2):915--931, 2011.

\bibitem{fully3d2017}
Hongming Shan, Ge~Wang, Mannudeep~K Kalra, R~de~Souza, and Junping Zhang.
\newblock Enhancing transferability of features from pretrained deep neural
  networks for lung nodule classification.
\newblock In {\em The Proceedings of the 2017 International Conference on Fully
  Three-Dimensional Image Reconstruction in Radiology and Nuclear Medicine
  (Fully3D)}, pages 65--68, 2017.

\bibitem{2017multilevel3d}
Qi~Dou, Hao Chen, Lequan Yu, Jing Qin, and Pheng-Ann Heng.
\newblock Multilevel contextual 3{D} {CNN}s for false positive reduction in
  pulmonary nodule detection.
\newblock {\em IEEE Transactions on Biomedical Engineering}, 64(7):1558--1567,
  2017.

\bibitem{lei2020shape}
Yiming Lei, Yukun Tian, Hongming Shan, Junping Zhang, Ge~Wang, and Mannudeep~K
  Kalra.
\newblock Shape and margin-aware lung nodule classification in low-dose {CT}
  images via soft activation mapping.
\newblock {\em Medical Image Analysis}, 60:101628, 2020.

\bibitem{wu2019learning}
Botong Wu, Xinwei Sun, Lingjing Hu, and Yizhou Wang.
\newblock Learning with unsure data for medical image diagnosis.
\newblock In {\em Proceedings of the IEEE International Conference on Computer
  Vision (ICCV)}, pages 10590--10599, 2019.

\bibitem{gutierrez2015ordinal}
Pedro~Antonio Guti{\'e}rrez, Maria Perez-Ortiz, Javier Sanchez-Monedero,
  Francisco Fernandez-Navarro, and Cesar Hervas-Martinez.
\newblock Ordinal regression methods: survey and experimental study.
\newblock {\em IEEE Transactions on Knowledge and Data Engineering},
  28(1):127--146, 2015.

\bibitem{liu2018ordinal}
Xiaofeng Liu, Yang Zou, Yuhang Song, Chao Yang, Jane You, and
  BV~K~Vijaya~Kumar.
\newblock Ordinal regression with neuron stick-breaking for medical diagnosis.
\newblock In {\em Proceedings of the European Conference on Computer Vision
  (ECCV)}, pages 335--344, 2018.

\bibitem{beckham2017unimodal}
Christopher Beckham and Christopher Pal.
\newblock Unimodal probability distributions for deep ordinal classification.
\newblock In {\em Proceedings of the International Conference on Machine
  Learning (ICML)}, pages 411--419, 2017.

\bibitem{shen2018deep}
Wei Shen, Yilu Guo, Yan Wang, Kai Zhao, Bo~Wang, and Alan~L Yuille.
\newblock Deep regression forests for age estimation.
\newblock In {\em Proceedings of the IEEE Conference on Computer Vision and
  Pattern Recognition (CVPR)}, pages 2304--2313, 2018.

\bibitem{zhu2020deep}
Haiping Zhu, Yuheng Zhang, Hongming Shan, Lingfu Che, Xiaoyang Xu, Junping
  Zhang, Jianbo Shi, and Fei-Yue Wang.
\newblock Deep ordinal regression forests.
\newblock {\em arXiv preprint arXiv:2008.03077}, 2020.

\bibitem{shu2019meta}
Jun Shu, Qi~Xie, Lixuan Yi, Qian Zhao, Sanping Zhou, Zongben Xu, and Deyu Meng.
\newblock Meta-weight-net: Learning an explicit mapping for sample weighting.
\newblock In {\em Advances in Neural Information Processing Systems (NeurIPS)},
  pages 1919--1930, 2019.

\bibitem{ResNet}
Kaiming He, Xiangyu Zhang, Shaoqing Ren, and Jian Sun.
\newblock Deep residual learning for image recognition.
\newblock In {\em The IEEE Conference on Computer Vision and Pattern
  Recognition (CVPR)}, June 2016.

\bibitem{vgg}
Karen Simonyan and Andrew Zisserman.
\newblock Very deep convolutional networks for large-scale image recognition.
\newblock In {\em International Conference on Learning Representations (ICLR)},
  pages 1--14, 2015.

\bibitem{kingma2014adam}
Diederik~P Kingma and Jimmy Ba.
\newblock Adam: A method for stochastic optimization.
\newblock {\em arXiv preprint arXiv:1412.6980}, 2014.

\end{thebibliography}
\end{document}